\documentclass[sigconf, nonacm]{acmart}
\usepackage{xcolor}
\usepackage{multirow}
\usepackage[acronym]{glossaries}
\usepackage{enumitem}
\usepackage{graphicx}
\usepackage[caption=true]{subfig}
\newacronym{cnns}{CNNs}{Convolutional Neural Networks}
\newacronym{nlp}{NLP}{Natural Language Processing}
\newacronym{vit}{ViT}{Vision Transformer}
\newacronym{sra}{SRA}{Spatial-Reduction Attention}
\newcommand{\mycirc}[1][black]{\textcolor{#1}{\ensuremath\bullet}}
\usepackage{color, colortbl}
\definecolor{Gray}{gray}{0.9}
\usepackage{cleveref}
\usepackage{graphicx}
\usepackage{caption}
\usepackage{afterpage}
\usepackage{indentfirst}
\begin{document}

\title{Cross-attention Spatio-temporal Context Transformer for Semantic Segmentation of Historical Maps}
\author{Sidi Wu}
\authornote{Both authors contributed equally to this research.}
\email{sidiwu@ethz.ch}
\affiliation{%
  \institution{Institute of Cartography and Geoinformation, ETH Zurich}
  \country{Switzerland}
}

\author{Yizi Chen}
\authornotemark[1]
\email{yizi.chen@ethz.ch}
\affiliation{%
  \institution{Institute of Cartography and Geoinformation, ETH Zurich}
  \country{Switzerland}
}

\author{Konrad Schindler}
\email{schindler@ethz.ch}
\affiliation{%
  \institution{Photogrammetry and Remote Sensing, ETH Zurich}
  \country{Switzerland}
  }

\author{Lorenz Hurni}
\email{lhurni@ethz.ch}
\affiliation{%
 \institution{Institute of Cartography and Geoinformation, ETH Zurich}
 \country{Switzerland}}

\renewcommand{\shortauthors}{Wu et al.}

\begin{abstract}
Historical maps provide useful spatio-temporal information on the Earth's surface before modern earth observation techniques came into being.
To extract information from maps, neural networks, which gain wide popularity in recent years, have replaced hand-crafted map processing methods and tedious manual labor.
However, aleatoric uncertainty, known as data-dependent uncertainty, inherent in the drawing/scanning/fading defects of the original map sheets and inadequate contexts when cropping maps into small tiles considering the memory limits of the training process, challenges the model to make correct predictions.
As aleatoric uncertainty cannot be reduced even with more training data collected, we argue that complementary spatio-temporal contexts can be helpful.
To achieve this, we propose a U-Net-based network that fuses spatio-temporal features with cross-attention transformers (U-SpaTem), aggregating information at a larger spatial range as well as through a temporal sequence of images.
Our model achieves a better performance than other state-or-art models that use either temporal or spatial contexts. 
Compared with pure vision transformers, our model is more lightweight and effective.
To the best of our knowledge, leveraging both spatial and temporal contexts have been rarely explored before in the segmentation task.
Even though our application is on segmenting historical maps, we believe that the method can be transferred into other fields with similar problems like temporal sequences of satellite images.
Our code is freely accessible at \url{https://github.com/chenyizi086/wu.2023.sigspatial.git}.
\end{abstract}


\maketitle
\begin{sloppypar}
\section{Introduction}
\begin{figure}[tb]
\centering
\subfloat[]{\includegraphics[width=0.32\linewidth]{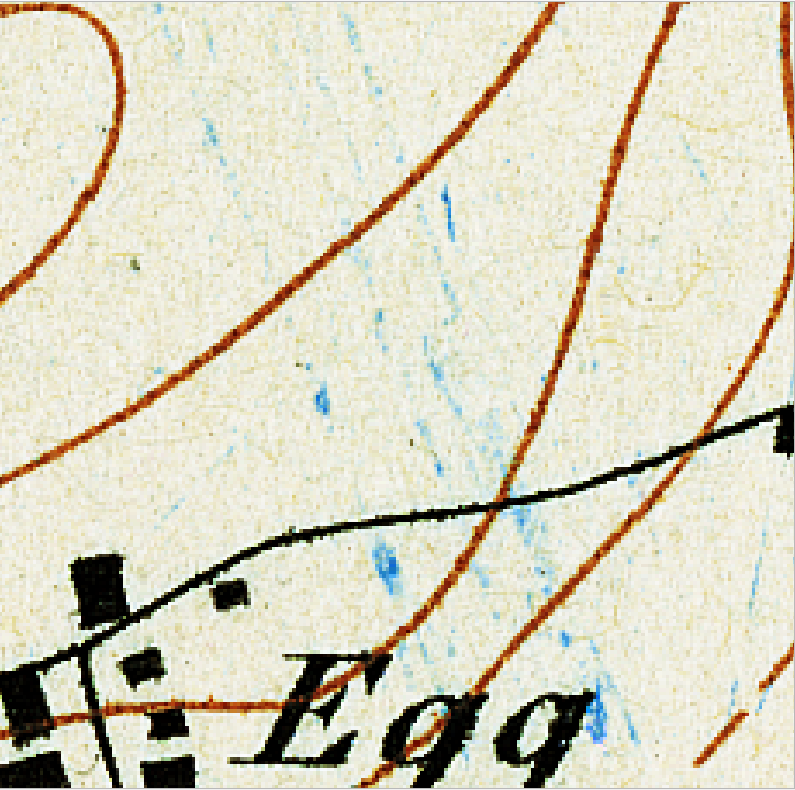}\label{fig:1a}}\hfill
\subfloat[]{\includegraphics[width=0.32\linewidth]{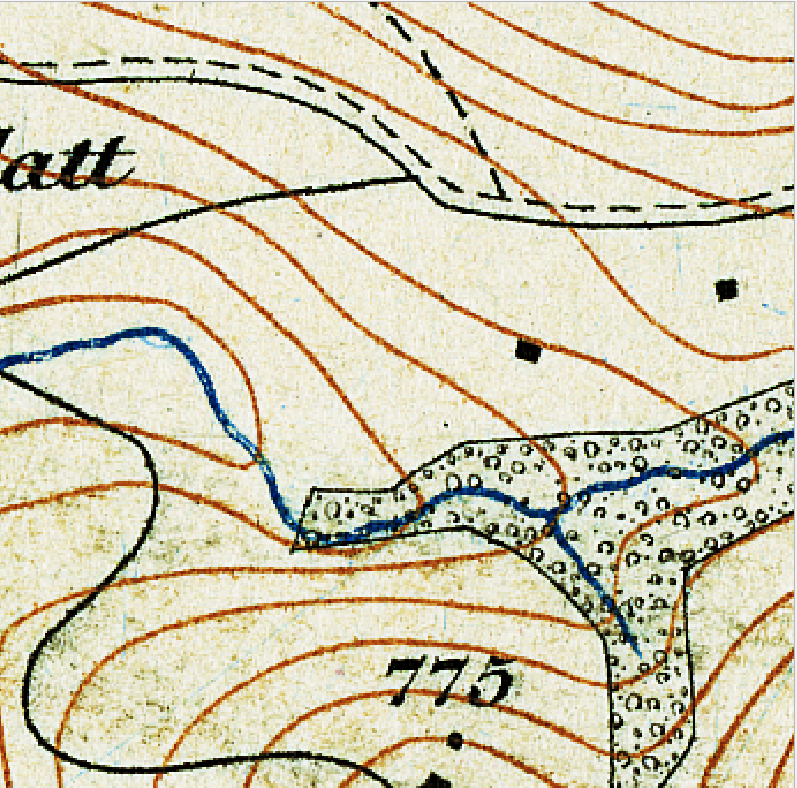}\label{fig:1b}}\hfill
\subfloat[]{\includegraphics[width=0.32\linewidth]{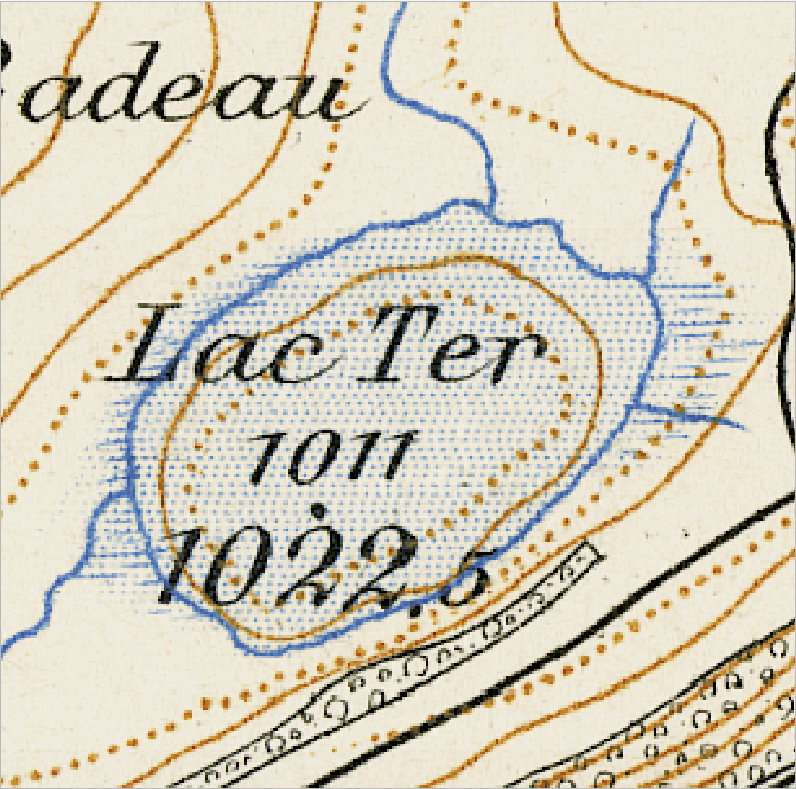}\label{fig:1c}}
\caption{Noise inherent in printing/scanning artifacts \protect\subref{fig:1a}, defects of painting strokes \protect\subref{fig:1b} and printing errors \protect\subref{fig:1c} where lake texture (dense dots) spreads out and mixes with wetland texture (short strokes).}
\label{fig:mapnoise}
\end{figure}

\begin{figure}[tb]
\centering
\includegraphics[width=\linewidth]{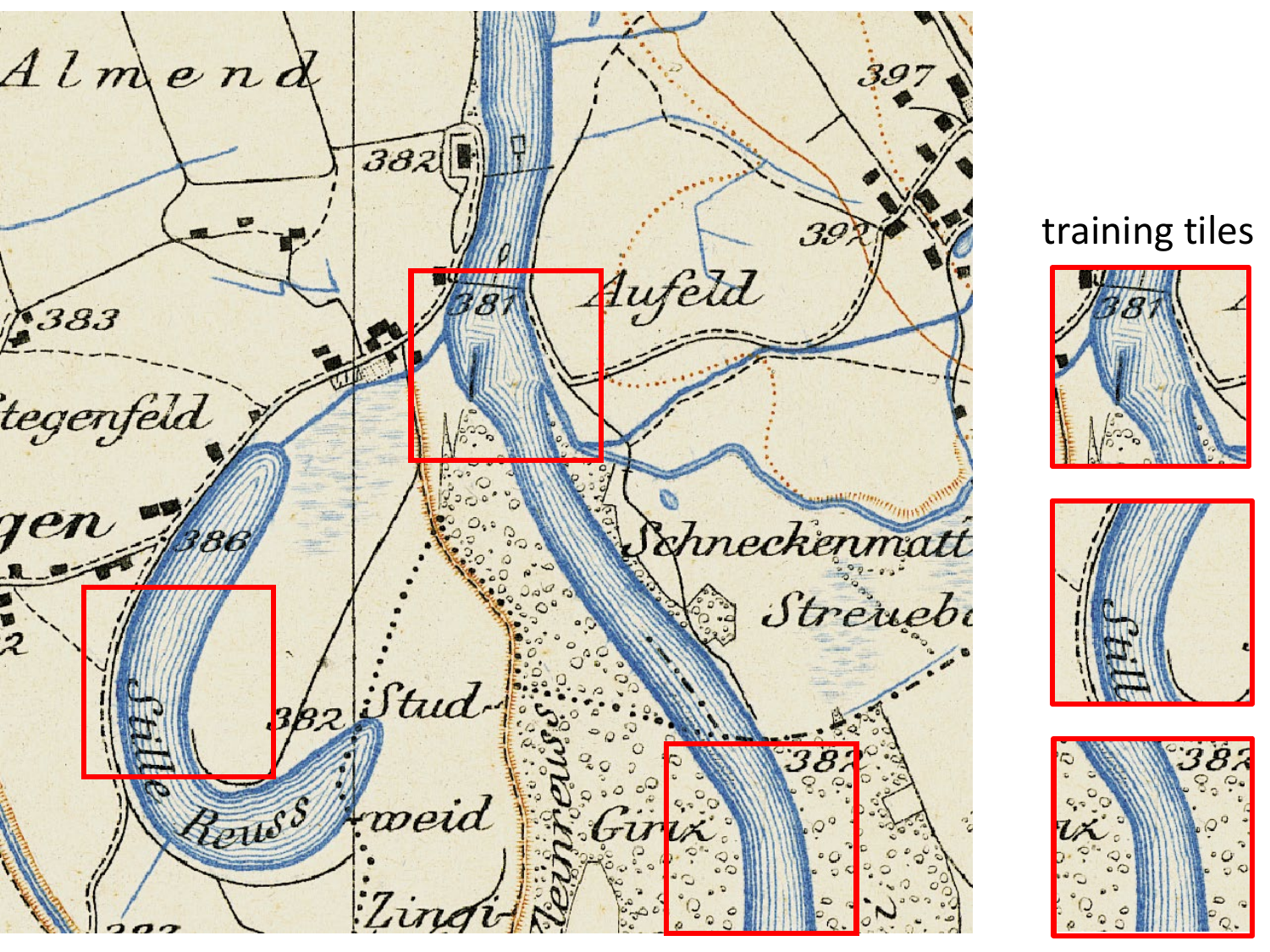}
\caption{Limited contexts when cropping an entire map sheet into small training tiles. Within these contexts distinguishing objects with similar symbols like a lake (left on the big image) and a river (right on the big image), is nearly impossible.}
\label{fig:riverandlake}
\end{figure}

Before the invention of modern air- and space-borne earth observation techniques, historical maps are almost the exclusive and comprehensive source to depict the spatio-temporal characteristics of the Earth's surface.
The topographic maps made two centuries ago can already have plausible accuracy thanks to the advances of surveying methods on a large scale.
They can be used to study the past states or changes of the Earth's surface \cite{Burghardt2022road, levin2010maps, san2014urban}.
Instead of manually retrieving information, which is time-consuming and tedious, recently \acrfull{cnns} have been favored to automate and accelerate the extraction process \cite{Heitzler2020cartographic, wu2022leveraging}.
However, extracting information from historical maps have inevitably aleatoric uncertainty \cite{kendall2017uncertainties, wu2022closer, wu2022leveraging}, known as data-dependent uncertainty, which comes from mainly two folds:
\begin{itemize}
    \item noise inherent in scanning artifacts, drawing and printing defects as well as fading papers of the map sheets, illustrated in \Cref{fig:mapnoise};
    \item insufficient context when cropping an entire map sheet into small training tiles (commonly used for deep learning), within which distinguishing objects with similar symbols is nearly impossible, shown in \Cref{fig:riverandlake}.
\end{itemize}
Implicated by its definition, aleatoric uncertainty cannot be reduced even with more training data collected.
One possible way to alleviate the aleatoric uncertainty is to add extra spatio-temporal contexts -- a larger temporal context is likely to include noise-less instances of the same object and a larger spatial context can provide a longer range of information to better differentiate objects.
Unlocking a larger spatio-temporal context is feasible because it's a general practice for cartographers to update maps regularly and usually the whole map was scanned and stored in the digital archive.
\\
A considerable amount of work has been done to incorporate spatial contexts into vision tasks \cite{chen2018deeplabv3, ding2018context, he2019adaptive, jin2021mining, li2020spatial, liu2020learning, yuan2020object}.
While these works explore contextual modules within the input image,  works like \cite{ding2022wide, Schmitz2021tissue, Yan2021hierachical} integrate larger contextual information outside the input image for applications like whole slide images (WSI) or high-resolution satellite images.
However, the straightforward down-sampling strategy for the contextual images in these works might have aliasing effects for maps with symbolic representations rather than natural images.
Temporal contexts have been widely researched in video segmentation \cite{jin2017video,liu2020efficient, nilsson2018semantic} and classification/segmentation of satellite image time series \cite{garnot2021panoptic, garnot2020satellite, russwurm2018convolutional}. 
While either spatial- or temporal- contexts have been well studied for the vision tasks, the benefits of incorporating both have been rarely explored. 
\\
To aggregate information from a larger spatial context, one might simply enlarge the input size to \acrshort{cnns}, which, however, will increase the memory consumption and does not consider the long-range dependency between pixels, due to the intrinsic locality of the convolution.
Transformers, which gained wide interest nowadays, have the powerful ability to model global relationships within all the input tokens based on the self-attention mechanism. 
They have been widely adopted from \acrfull{nlp} to computer vision \cite{dosovitskiy2020vit,liu2021swin,wang2021pvt} as an alternative to \acrshort{cnns}.
However, applying transformers directly on the input image as well as its spatio-temporal contextual images is computationally expensive, considering the total number of pixels. 
Some works \cite{ding2022wide, garnot2021panoptic, liu2021transformer} take advantage of both \acrshort{cnns} and transformers -- the former for better feature down-sampling and representation and the latter for long-range dependencies. 
Our work follows this direction of research.
Besides spatial contexts, our model also aggregates temporal contexts in one go.
Different from existing approaches that aggregate temporal information at the same location \cite{cciccek20163dUNET, garnot2021panoptic}, our method relates all the pixels between temporally-neighboring samples.
This accounts for the shift of geographical objects at different years due to surveying/map-making/geo-referencing bias on the one hand and integrates additional spatial information of temporal neighbors on the other hand.
\\
Our main contributions can be summarized as follows:
\begin{itemize}
    \item We integrate spatial and temporal contexts simultaneously for better historical map segmentation in the presence of aleatoric uncertainty.
    \item Our proposed simple yet effective network U-SpaTem takes advantage of both CNNs and transformers with a cross-attention mechanism instead of the commonly-used self-attention mechanism.
\end{itemize}

\section{Related Work}
\subsection{Historical Map Segmentation}
To unlock the spatial information of the Earth's surface in the past and track the long-term evolution of natural and man-made objects, historical maps serve as useful sources.
Tremendous scanned historical maps stored in digital archives have sparked automatic map processing methods in the past two decades \cite{Bin1998, chiang2009extracting, leyk2009segmentation}.

With the advances of neural networks for the computer vision tasks,  automatic map processing has shifted towards \acrshort{cnns}-based methods \cite{chen2016attention, chen2018deeplabv3, ronneberger2015u, visin2016reseg}, for extracting buildings \cite{Heitzler2020cartographic, Uhl2020building}, building blocks \cite{chen2021combining, jianhua2021hatch}, roads \cite{ekim2021road}, water bodies \cite{wu2022leveraging, WU2023DA}, and texts \cite{arundel2022deep, li2021synthetic}.

\subsection{Transformer in Vision Tasks}
Firstly introduced for \acrshort{nlp} \cite{vaswani2017attention}, transformers account for long-range dependencies of sequence data based on the self-attention mechanism within all the input tokens. With this mechanism, the model learns to focus on the most salient tokens.
The \acrfull{vit}, proposed by \cite{dosovitskiy2020vit}, applies the transformer to images by splitting images into patches and treating feature vectors of these patches as tokens.
Despite the state-of-art performance of transformers in segmentation \cite{xie2021segformer, zheng2021rethinking}, the use of transformers in a large spatial-temporal context is non-trivial if we regard each patch from the target image as well as from the spatio-temporal contextual images as a token and apply self-attention to all tokens.
To reduce the computational complexity, \cite{sun2022coarse} proposed a cross-attention mechanism between the central feature and its contexts.
Works like \cite{ding2022wide, garnot2021panoptic, liu2021transformer} combine transformers and \acrshort{cnns}, which not only improves the training efficiency but also takes both advantages of CNNs for locality modeling and transformers for global dependencies, arguably both essential for vision tasks \cite{liu2021transformer}.
Our work follows this line of research and extends the work of \cite{garnot2021panoptic} for integrating both spatio-temporal contexts with a cross-attention mechanism.

\section{Methodology}
\subsection{Network Architecture}
\begin{figure*}[ht]
    \centering
    \includegraphics[width=0.9\textwidth]{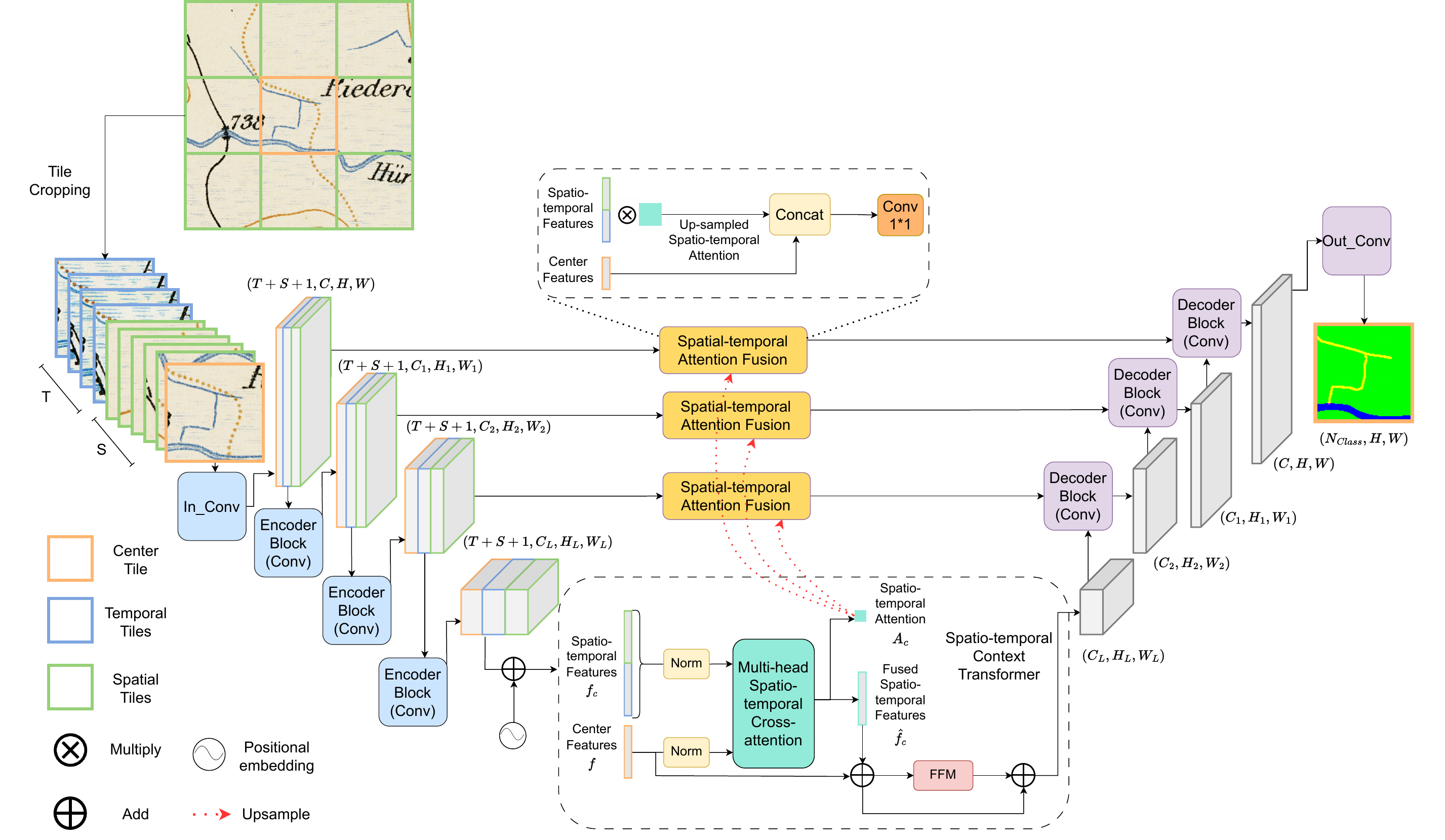}
    \caption{The proposed architecture (U-SpaTem).}
    \label{fig:architectueroverall}
\end{figure*}

Our proposed architecture U-SpaTem (U-Net with Spatio-Temporal Context Transformer) is illustrated in \Cref{fig:architectueroverall}.
Given a map tile, we get its adjacent tiles in the same map as spatial contexts and tiles at the same location from its temporally consecutive maps as temporal contexts.
The central tile, spatial tiles, as well as temporal tiles, are processed simultaneously and independently by the encoder blocks, each of which is made up of down-sampling (strided convolution) and a sequence of convolutions, ReLU activation, and batch normalization, to gradually reduce the resolution of the input images.
The \texttt{In\_Conv} in the first level differs from other encoder blocks in that no down-sampling is used.
No cross-tile convolution is used like 3D U-Net \cite{cciccek20163dUNET}.\\
The bottleneck features at the end of encoder blocks are passed through a Spatio-Temporal Context Transformer, to get the fused features weighted by the spatio-temporal attention masks. The masks will then be upsampled to fuse spatio-temporal features at different depths. 
The details will be introduced later.
The fused bottleneck features are then passed to the decoder blocks, each of which is made up of up-sampling (strided transposed convolution) and a sequence of convolutions, ReLU activation, and batch normalization.
The fused spatio-temporal features from the encoder are passed to the decoder block at the same resolution through skip connections.
The \texttt{Out\_Conv} differs from other decoder blocks as it's a single convolution operation to map feature channels to the number of classes, followed by Sigmoid activation to generate a per-class segmentation map for the central tile.

\subsection{Spatio-temporal Context Transformer}
The bottleneck features at the end of the encoder are processed independently for each temporal and spatial tile through one multi-head spatio-temporal cross-attention block. 
Instead of adding an attention block at each depth, which will increase the memory drastically, we only conduct cross-attention on bottleneck features and up-sample the attention masks to all resolutions, to fuse spatio-temporal features at different depths, following \cite{garnot2021panoptic}.
We apply layer normalization(LN) \cite{xiong2020layer} to all the features before the attention blocks.
\\
For transformers, it is important to let the network be aware of the sequence order by adding positional encoding to inject relative or absolution positions into tokens of a sentence \cite{vaswani2017attention}.
Instead of hard-set positional encoding, we add to the features the learnable positional embeddings as \cite{fan2021msvit} before the attention blocks.

\subsubsection{\textbf{Multi-head Cross-attention}}
Different from the self-attention mechanism, where the query, value, and key come from a single embedding sequence, cross-attention combines two separate embedding sequences, where the key and value come from one sequence and the query comes from the other.
Let $t_{1}\in \mathbb{R}^{N \times D}$ and $t_{2}\in \mathbb{R}^{M \times D}$ denote two separate embedding sequences, where $D$ denotes the feature dimensions, and $M, N$ denote the lengths of the sequences, respectively. 
Then the query $\mathbf{Q}$ is derived from $t_{1}$ and the key $\mathbf{K}$ and value $\mathbf{V}$ are from $t_{2}$:
\begin{equation}
\mathbf{Q} = t_{1}\mathbf{W}^{Q}, \mathbf{K} = t_{2}\mathbf{W}^{K}, \mathbf{V} = t_{2}\mathbf{W}^{V},
\end{equation}
where the $\mathbf{W}^{Q} \in \mathbb{R}^{D \times D}$, $\mathbf{W}^{K} \in \mathbb{R}^{D \times D}$ and $\mathbf{W}^{V} \in \mathbb{R}^{D \times D}$ are parameters for linear projection. 
Then we have the attention $\mathbf{A} = Softmax(\frac{QK^{T}}{\sqrt{D}}) \in \mathbb{R}^{N \times M}$, and $t_{1}$ will be updated by:
\begin{equation}
    \hat{t}_{1} = A V \mathbf{W}^P + t_{1},
\end{equation}
with the residual connection. $\mathbf{W}^P \in \mathbb{R}^{D \times D}$ is a fully connected layer for feature projection. We use multi-head attention with $n$ heads, which divides features into $n$ groups along the channel dimension $D$, and thus runs $n$ multiple cross-attention in parallel. Each group $g$ has then the channel dimension of $D_{head} = \frac{D}{n}$.
Each attention can be denoted as $A_{g} = Softmax(\frac{Q_{g}K_{g}^{T}}{\sqrt{D_{head}}})$ where $Q_{g} \in \mathbb{R}^{N \times D_{head}}$ and $K_{g} \in \mathbb{R}^{M \times D_{head}}$. We then concatenate features obtained from each attention head along the channel dimension:
\begin{equation}
    \hat{t}_{1} = Concat(A_{g}V_{g})^n_{g=1}\mathbf{W}^P+ t_{1}, 
    \label{eq:multihead}
\end{equation}
where $V_{g} \in \mathbb{R}^{M \times D_{head}}$. 
\\
The output $\hat{t}_{1} \in \mathbb{R}^{N \times D}$ is passed to a feed-forward module $FFM$ consisting of LN and multi-layer perceptron (MLP), with the residual connection:
\begin{equation}
\begin{aligned}
FFM(\hat{t}_{1}) = MLP(LN(\hat{t}_{1})),\\
\hat{t}_{1} = FFM(\hat{t}_{1}) + \hat{t}_{1}.
\end{aligned}
\end{equation}

\subsubsection{\textbf{Multi-head Spatio-Temporal Cross-attention}}\label{sec:spacrossatt}
\begin{figure}[tb]
    \centering
    \includegraphics[width=\linewidth]{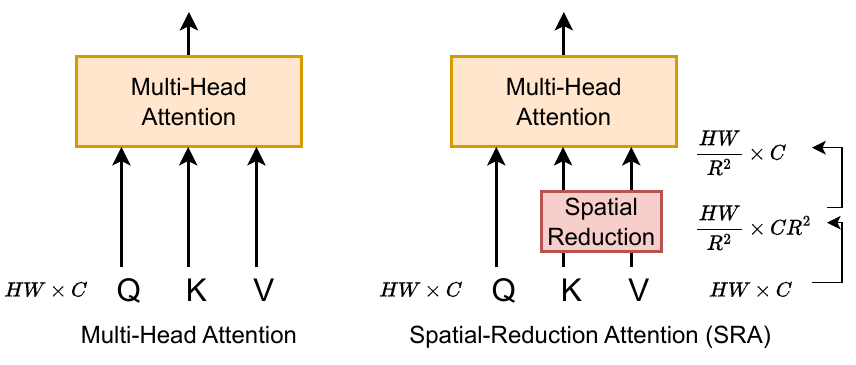}
    \caption{Spatial reduction attention (SRA) \cite{wang2021pvt}.}
    \label{fig:SRA}
\end{figure}

We use cross-attention between the central feature $f \in \mathbb{R}^{H_L \times W_L \times D}$  and its contextual features $f_{c} \in$ $\mathbb{R}^{I \times H_L \times W_L \times D}$, including $S$ spatial tiles and $T$ temporal tiles, in total $I = S + T$ tiles,  to fuse the spatio-temporal contexts at the depth $L$.
The multi-head cross-attention is carried out between the central tile and each contextual tile $i$ independently with shared weights $\mathbf{W}^{Q}_{c} \in \mathbb{R}^{D \times D}, \mathbf{W}^{K}_{c} \in \mathbb{R}^{D \times D}, \mathbf{W}^{V}_{c} \in \mathbb{R}^{D \times D}$: 
\begin{equation}
\begin{aligned}
Q_{c} = f{\mathbf{W}^{Q}_{c}}, 
K^{i}_{c} = f^{i}_{c}{\mathbf{W}^{K}_{c}},
V^{i}_{c} = f^{i}_{c}{\mathbf{W}^{V}_{c}};
i \in [1, \dots, I].
\end{aligned}
\end{equation}
To save memory cost of attention operation, we make use of spatial-reduction attention (\acrshort{sra}) proposed by \cite{wang2021pvt}, which uses a reduction ratio $R$ to reduce image resolution, illustrated in \Cref{fig:SRA}:
\begin{equation}
    SR(f^i_c)= Reshape(f^i_c,R)W^{SR}.
\end{equation}
$Reshape(f^i_c, R)$ is to reshape $f^i_c$ of shape $H_L \times W_L \times D$ to $\frac{H_L}{R} \times \frac{W_L}{R} \times (R^2D)$.
It is followed by a linear projection layer $W^{SR} \in \mathbb{R}^{(R^2D) \times D}$ to reduce the channel dimension from $(R^2D)$ to $D$.
Then we have:
\begin{equation}
 Q_{c} = f{\mathbf{W}^{Q}_{c}}, 
K^{i}_{c} = SR(f^{i}_{c}){\mathbf{W}^{K}_{c}},
V^{i}_{c} = SR(f^{i}_{c}){\mathbf{W}^{V}_{c}};
i \in [1, \dots, I]. 
\label{eq:SRattention}
\end{equation}
We derive an attention map for each contextual tile $i$: 
\begin{equation}
    A^i_{c} = Softmax(\frac{Q_{c}{K^i_{c}}^{T}}{\sqrt{D_{head}}}); A^i_{c} \in \mathbb{R}^{(H_L\times W_L) \times (h_L\times w_L)},
\label{eq:attention}
\end{equation}
with $h_L = \frac{H_L}{R}$ and $w_L = \frac{W_L}{R}$.
Here we omit the formulation of multi-head attention as in \Cref{eq:multihead} for simplicity.
We then have updated spatial and temporal features $\hat{f}^i_c$ weighted by the attention maps:
\begin{equation}
    \hat{f}^i_c =  A^i_{c}V^i_c\mathbf{W}^P_c; i \in [1, \dots, I].
    \label{eq:attentioneachtile}
\end{equation}
The computation complexity for \Cref{eq:attentioneachtile} is reduced by $R^2$ compared with no spatial reduction.

\subsubsection{\textbf{Spatio-temporal Feature Fusion}}
We concatenate $\hat{f}^i_c$ among I tiles to  the shape of $I \times H_L \times W_L \times D$.
We use a 1 $\times$ 1 convolution operation to weight and fuse $I$ spatial/temporal features to one spatio-temporal feature (attentive feature pooling).
We obtain the resulting feature:
\begin{equation}
    \hat{f}_c = Conv_{1\times1}(Concat(\hat{f}^i_c)^{I}_{i=1}); \hat{f}_c \in H_L \times W_L \times D.
\end{equation}
Then we update the center feature: 
\begin{equation}
    \hat{f} = f + \hat{f}_c.
\end{equation}
This is followed by Layer Normalization and a feed-forward module ($FFM$) with a residual connection.
The $FFM$ consists of two linear transformations with GeLU \cite{hendrycks2016gaussian} activation and dropouts in between as in \cite{wang2021pvt}:
\begin{equation}
    \hat{f} = FFM(\hat{f}) + \hat{f}.
\end{equation}

\begin{table*}[t]
\caption{Comparison with other methods. We report mean Intersection over Union (mIoU), mean F1 and per-class F1 as well as parameters (million(M)) and inference time (for 100 samples); \mycirc[green]: U-SpaTem uses spatial context as input; \mycirc[red]: U-SpaTem uses temporal context as input; (\mycirc[green], \mycirc[red]): U-SpaTem uses both spatial and temporal context as input. As \cite{wu2022leveraging} was implemented in Keras \cite{chollet2015keras} while others were in Pytorch \cite{paszke2017automatic}, we omit parameters and inference time for \cite{wu2022leveraging} for comparison.}
\begin{tabular}{lcccccccccc}
\toprule
\multirow{2}{*}{Method} & \multirow{2}{*}{\shortstack{Spatial\\ context}} & \multirow{2}{*}{\shortstack{Temporal\\ context}} & \multicolumn{4}{c}{Per-class F1(\%) $\uparrow$ } & \multirow{2}{*}{mean F1(\%) $\uparrow$ } & \multirow{2}{*}{mIoU(\%) $\uparrow$} & \multirow{2}{*}{Params(M) $\downarrow$} & \multirow{2}{*}{IT(s) $\downarrow$} \\ \cline{4-7}
& & & Stream & Wetland & River & Lake & & & & \\ 
\midrule
U-Net~\cite{ronneberger2015u}   & - & - & \textbf{96.44} & 88.30 & 64.22 & 51.80 & 75.19 & 74.34 & 1.08M & 0.97s  \\
\rowcolor{Gray}
3D U-Net~\cite{cciccek20163dUNET}  & - & \checkmark   & 95.55 & 88.54 & 64.58 & 54.37 & 75.76 & 74.63 & 1.54M & 1.94s \\
U-TAE~\cite{garnot2021panoptic}    & - & \checkmark & 89.40 & 88.41 & 62.98 & 55.80 & 74.15 & 73.08 &  1.06M & 2.08s  \\
\rowcolor{Gray}
U-ASPP-LCE~\cite{wu2022leveraging}  & - & - & 96.35 & 88.50 & 65.49 & 53.65 & 76.00 & 75.12 & -- & -- \\
Segformer~\cite{xie2021segformer}  & \checkmark & - & 87.72 & 68.16 & 41.65 & 58.79 & 64.08 & 61.40 & 3.72M & 1.63s \\
\midrule
\rowcolor{Gray}
U-SpaTem (Ours) \mycirc[green] & \checkmark & - & 94.95 & 89.49 & 65.47 & \textbf{65.15} & \textbf{78.77} & 77.55 & 1.98M & 1.63s\\
U-SpaTem (Ours) \mycirc[red]              & - & \checkmark & 96.20 & 90.82 & \textbf{65.68} & 62.24 & 78.74 & \textbf{77.84} & 1.98M & 1.19s \\
\rowcolor{Gray}
U-SpaTem (Ours) (\mycirc[green], \mycirc[red]) & \checkmark & \checkmark & 94.93 & \textbf{91.31} & 64.61
 & 63.69 & 78.64 & 77.54 & 2.44M & 1.94s \\
\bottomrule
\end{tabular}
\label{tab:experiment comparison}
\end{table*}

\subsection{Aggregating Spatio-temporal Attention At All Depths}
Similar to \cite{garnot2021panoptic}, we up-sample spatial and temporal attention masks and use them to fuse spatio-temporal features at different depths.
\\
We concatenate $A^i_c$ from \Cref{eq:attention} among $I$ tiles and we have $A^L_c$ of the size $ H_L \times W_L \times (I \times h_L\times h_L)$ at the depth $L$. 
For each depth $l = 1, ..., L$ of the encoder, we up-sample the attention map $A^L_c$ to $A^l_c$ of the size $H_l \times W_l \times (I \times h_L\times h_L)$, respectively.
\\
For spatial/temporal features $f^l_c$ of the shape $I \times H_l \times W_l \times C_l$, we first down-sample them into  $ (I \times h_L \times w_L) \times C_l$ using max pooling and multiply them with $A^l_c$ to obtain the fused spatio-temporal contexts of the size $H_l \times W_l \times C_l$ at depth $l$.
Since we have $n$ attention maps out of $n$ heads used in the transformer, we split $f^l_c$ into $n$ groups along the channel dimension $C_l$ and the resulting maps are concatenated back along the channel dimension:
\begin{equation}
    f^l_{c} = Concat(A^{l,g}_cf^{l,g}_c)^{n}_{g=1}.
\end{equation}
\\
We combine the central features with the spatio-temporal contextual features derived above to get the fused features at depth $l$:
\begin{equation}
    \hat{f}^l = Conv_{1 \times 1}(Concat(f^l, f^l_c)).
\end{equation}
with a $1 \times 1$ convolution layer of $C_l$ channels. $\hat{f}^l$ is passed to the decoder at the same depth (skip connection).

\begin{figure*}[tb]
  \centering
  \setlength\tabcolsep{1pt}
  \begin{tabular}{cccccccc}
    Input & Annotation & U-Net & 3D U-Net & U-TAE & U-ASPP-LCE & Segformer & U-SpaTem (ours) \\
    \includegraphics[width=0.12\textwidth]{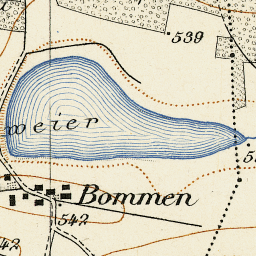}&
    \includegraphics[width=0.12\textwidth]{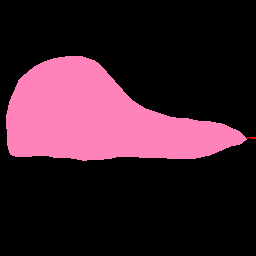}&
    \includegraphics[width=0.12\textwidth]{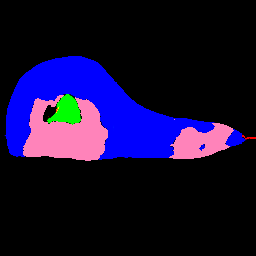}&
    \includegraphics[width=0.12\textwidth]{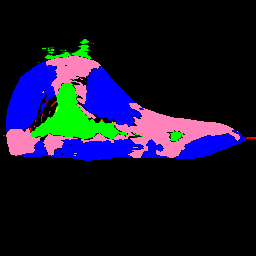}&
    \includegraphics[width=0.12\textwidth]{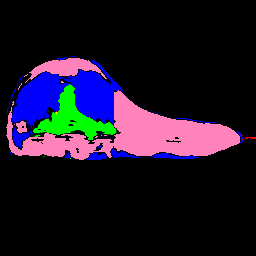}&
    \includegraphics[width=0.12\textwidth]{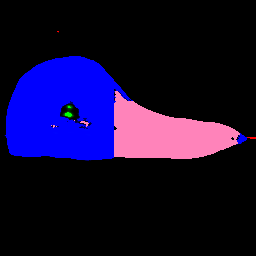}&
    \includegraphics[width=0.12\textwidth]{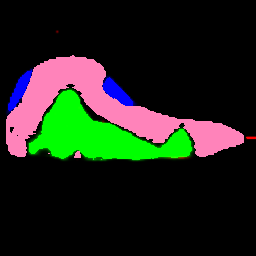}&
    \includegraphics[width=0.12\textwidth]{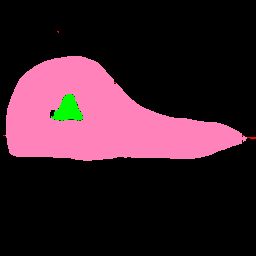}\\
    \includegraphics[width=0.12\textwidth]{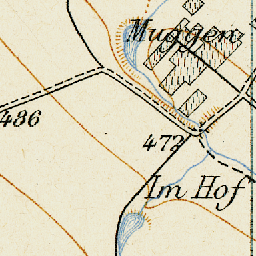}&
    \includegraphics[width=0.12\textwidth]{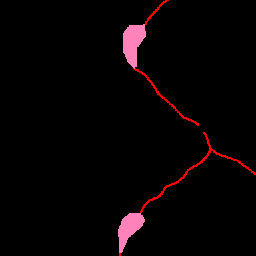}&
    \includegraphics[width=0.12\textwidth]{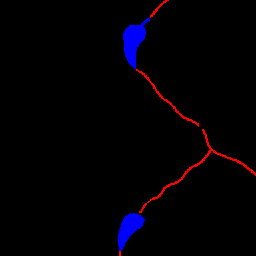}&
    \includegraphics[width=0.12\textwidth]{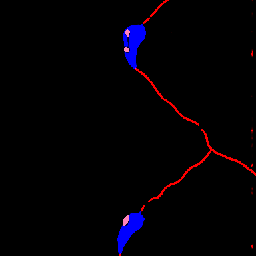}&
    \includegraphics[width=0.12\textwidth]{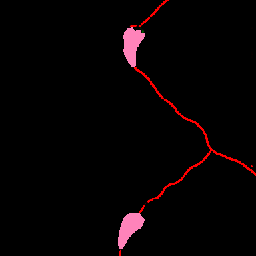}&
    \includegraphics[width=0.12\textwidth]{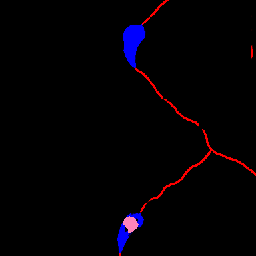}&
    \includegraphics[width=0.12\textwidth]{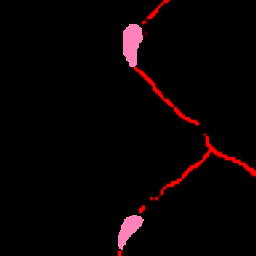}&
    \includegraphics[width=0.12\textwidth]{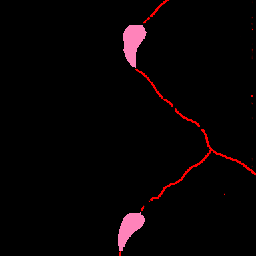}\\
    \includegraphics[width=0.12\textwidth]{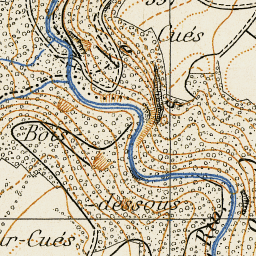}&
    \includegraphics[width=0.12\textwidth]{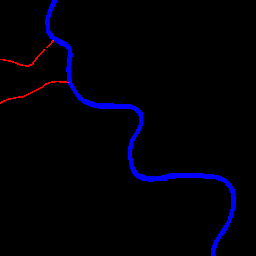}&
    \includegraphics[width=0.12\textwidth]{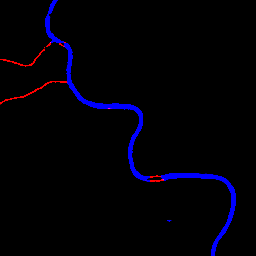}&
    \includegraphics[width=0.12\textwidth]{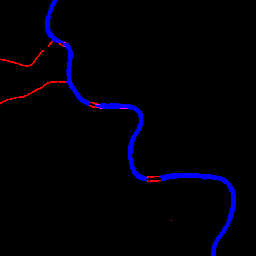}&
    \includegraphics[width=0.12\textwidth]{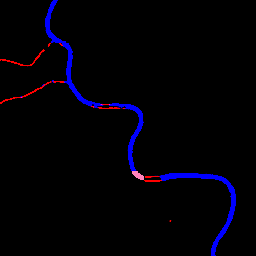}&
    \includegraphics[width=0.12\textwidth]{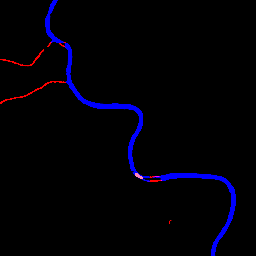}&
    \includegraphics[width=0.12\textwidth]{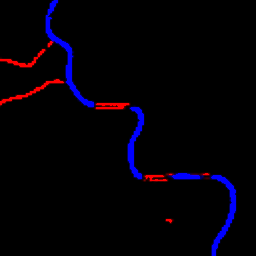}&
    \includegraphics[width=0.12\textwidth]{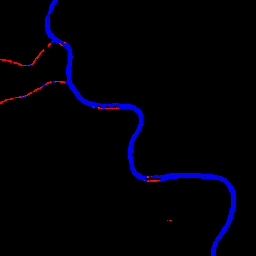}\\
    \includegraphics[width=0.12\textwidth]{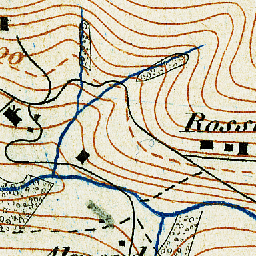}&
    \includegraphics[width=0.12\textwidth]{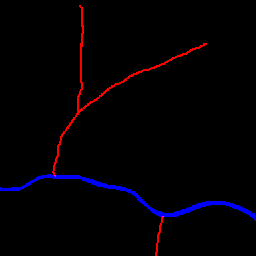}&
    \includegraphics[width=0.12\textwidth]{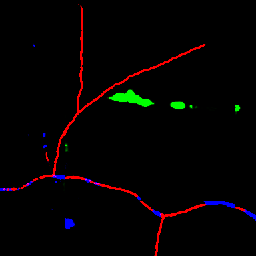}&
    \includegraphics[width=0.12\textwidth]{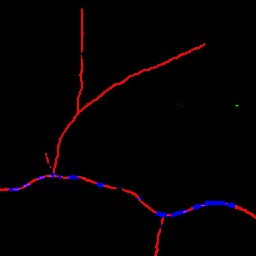}&
    \includegraphics[width=0.12\textwidth]{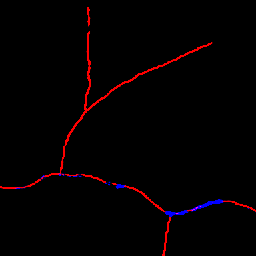}&
    \includegraphics[width=0.12\textwidth]{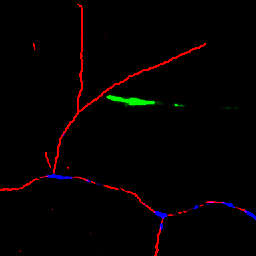}&
    \includegraphics[width=0.12\textwidth]{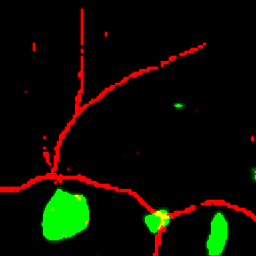}&
    \includegraphics[width=0.12\textwidth]{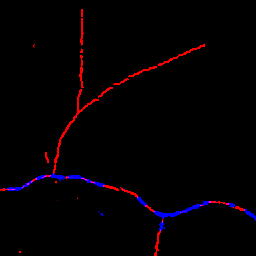}\\
  \end{tabular}
  \caption{Qualitative comparison between different methods. In both predictions and annotations, red, green, blue, and pink represent streams, wetlands, river, and lakes, respectively.}
  \label{fig:figure_qualitative}
\end{figure*}

\section{Experiments and Results}
\subsection{Data Preparation}
We make use of Siegfried maps\footnote{\url{https://www.swisstopo.admin.ch/en/geodata/maps/historical/siegfried25.html}}, a Swiss national topographic map series that was published between 1870 and 1949, at the scale of 1:25000 for the Swiss plateau and 1:50000 for the Swiss Alps.
The map sheets were scanned and archived by the Federal Office of Topography (Swisstopo).
We used 60 maps and their consecutive temporal sheets at the scale of 1:25000 around the year 1880 in our experiment.
Each map sheet is 7000 $\times$ 4800 pixels with a spatial resolution of 1.25m and has the ground-truth annotation maps for streams, wetlands, rivers, and lakes. 
We split the map sheets into training (40), validation (10), and testing (10).
For training and validation, we randomly sample tiles of 256 $\times$ 256 pixels and their spatio-temporal contextual tiles of the same size, for each class equally. We empirically fix the ratio between the number of positive samples (containing objects of interest) and negative samples (not containing objects of interest) as 4:1.
In total, we have 2793 sequences for training and 671 sequences for validation. 
Each sequence has a length of 13, including one central tile, four temporally-consecutive tiles at the same location, and eight spatially-adjacent tiles in the 3 $\times$ 3 neighborhood.
We evaluate our model on the entire 10 testing sheets.

\subsection{Implementation Details}
Our U-SpaTem has $L = 5$ depths. 
We use $n=16$ heads for our Spatio-temporal Context Transformer and $R=4$ for SRA.
For our experiments, we use Adam optimizer \cite{kingma2014adamold} with an initial learning rate of 0.001 and a decay of 0.5 if the validation loss does not decrease for 5 epochs, until 0.00001. 
We use dice loss for training, considering the feature sparsity.
The batch size in all our experiments is fixed as 32 and we train each model for 50 epochs.
We compare our model with other promising models:
\begin{itemize}
    \item U-Net \cite{ronneberger2015u}, which is our baseline model without introducing any spatial/temporal contexts.
    \item 3D U-Net \cite{cciccek20163dUNET}, an extended U-Net model with 3D convolutions in both encoder and decoder blocks to additionally aggregate temporal information.
    \item U-TAE \cite{garnot2021panoptic}, a U-Net-based model with a transformer between the encoder and decoder blocks to fuse the temporal features with attention.
    \item U-ASPP-LCE \cite{wu2022leveraging}, a state-of-art U-Net model for hydrology segmentation in historical maps with Atrous Spatial Pyramid Pooling (ASPP) and uncertainty estimation.
    \item Segformer \cite{xie2021segformer}, a state-of-art transformer-based segmentation model.
\end{itemize} 
For Segformer \cite{xie2021segformer}, we initialize the network with the provided pre-trained weights.
An image of $768 \times 768$ is input directly to the Segformer so that the equivalent spatial context as our model is included. 
By default, the image will be reshaped to 512 before the transformer blocks. 
Finally, the output of $128 \times 128$ is up-sampled to restore the original resolution.

\subsection{Results and Discussion}
\begin{figure*}[tb]
  \resizebox{0.44\textheight}{!}{%
    \setlength\tabcolsep{1pt}
    \begin{tabular}{cc}
      \includegraphics[width=0.48\linewidth]{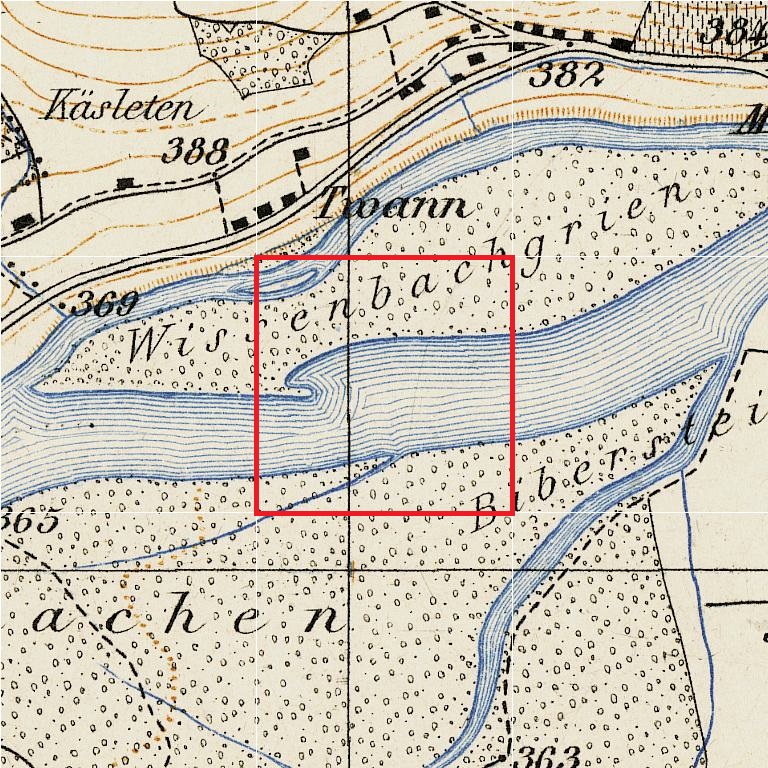} &
      \includegraphics[width=0.48\linewidth]{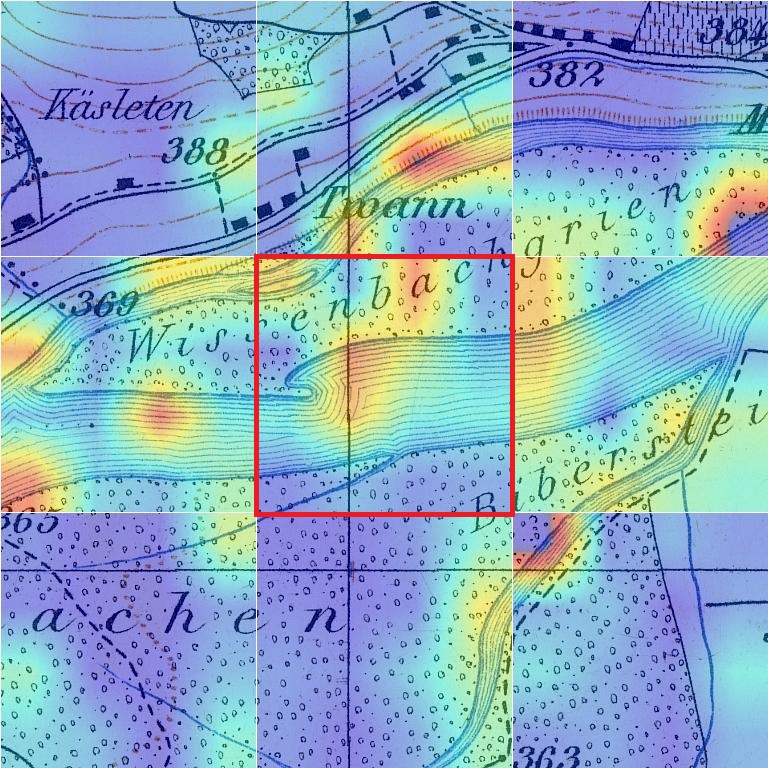} \\
      \includegraphics[width=0.48\linewidth]{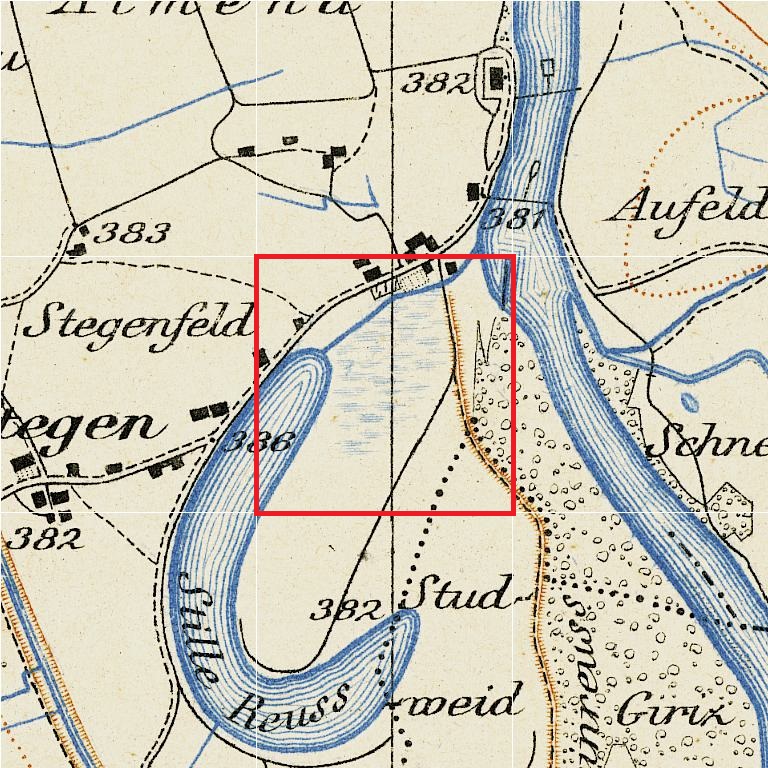} &
      \includegraphics[width=0.48\linewidth]{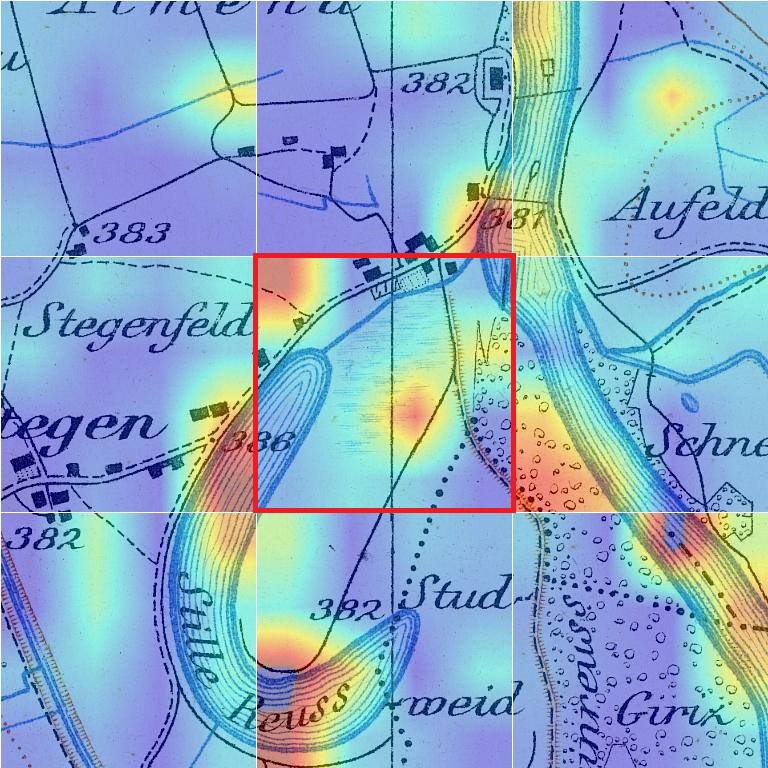} \\
    \end{tabular}
  }
  \caption{Visualization of spatial attention. Blue to red represents values from low to high. We visualize the maximum value among all the attention heads. The input tiles are on the left side, and the attention maps are on the right side. The central tile is marked as a red square.}
  \label{fig:spatialattention}
\end{figure*}

We compare the segmentation results of our proposed U-SpaTem and other methods, shown in \Cref{tab:experiment comparison} (the upper part).
We mark which type(s) of context each method utilizes. 
We conduct ablation studies on spatial and temporal contexts and our proposed architecture is equally applicable no matter if only one or both contexts are available.
The results are shown in \Cref{tab:experiment comparison} (the lower part).
\\
As we can see, adding spatial contexts largely helps to refine the predictions of lakes, to which contexts can often be vital.
Compared with Segformer, our spatial-context-only model has half the number of parameters and much higher accuracy.
Compared with U-TAE and 3D U-Net which fuses temporal information only at the same location, our temporal-context-only model also takes the spatial information of temporal neighbors into account.  
Our model has a slightly higher number of parameters but a much shorter inference time and better segmentation accuracy.
From our ablation studies, we see that adding temporal information improves the segmentation of wetlands that may have faded strokes/scanning artifacts occasionally.
While adding either temporal or spatial context leads to an obvious improvement for certain classes, fusing both spatial and temporal contexts does not yield a better result in our experiment.
\\
\Cref{fig:figure_qualitative} shows the qualitative comparison, for lakes in the first two examples, streams and rivers with unstable painting strokes in the last two examples, and scanning artifacts in the last example.
Our model has achieved a better overall performance in front of all these scenarios/challenges.
\\
We visualize spatial attention maps in \Cref{fig:spatialattention}, where the center tile (marked in red) has an inadequate context to differentiate between rivers and lakes that have the same symbolization, as illustrated in \Cref{fig:riverandlake}.
Higher values indicate more salient/important features the model pays attention to.
We see that the attention maps have strong activation on the neighboring tiles, which means that the model tries to collect information about the shape and spatial extent of the feature. 
In Figure 7, where the center tile has wetlands of faded strokes, the model pays attention to other temporal instances of the same wetland to gather temporal information, while ignoring the changes (in the last two images of the second example).

\section{Conclusion}
We have proposed a simple yet effective model U-SpaTem that leverages the advantages of both CNNs, which can represent features and preserve spatial information effectively, and transformers, which can aggregate information from a larger spatio-temporal context.
The additional spatio-temporal information is especially useful for segmenting historical maps where aleatoric uncertainty is present due to defects of original map sheets or limited contexts caused by tile cropping during training.
Our model shows a better performance compared with other models that use either temporal or spatial context.
Also, our model is more lightweight and effective than pure transformer-based models and does not need to sacrifice the resolution of predictions.
In our experiment, while temporal context is more important to wetlands that sometimes have faded strokes, spatial context is more vital to lakes that can have inadequate context due to tile cropping.
However, fusing both spatial and temporal contexts does not lead to an obvious improvement, which means that it might not be necessary to always involve both contexts in our case.
To find out which type(s) of context is most suitable for each class, it is worth investigating all different possibilities, where our proposed architecture is equally applicable. A better strategy for combining both contexts might be further investigated.
Despite that our use case is historical maps, the proposed model is general enough for other geospatial applications like segmenting satellite images given a temporal sequence.

\begin{figure*}[tb]
\resizebox{0.44\textheight}{!}{
\setlength\tabcolsep{1pt}
\begin{tabular}{cc}
\includegraphics[width=\linewidth]{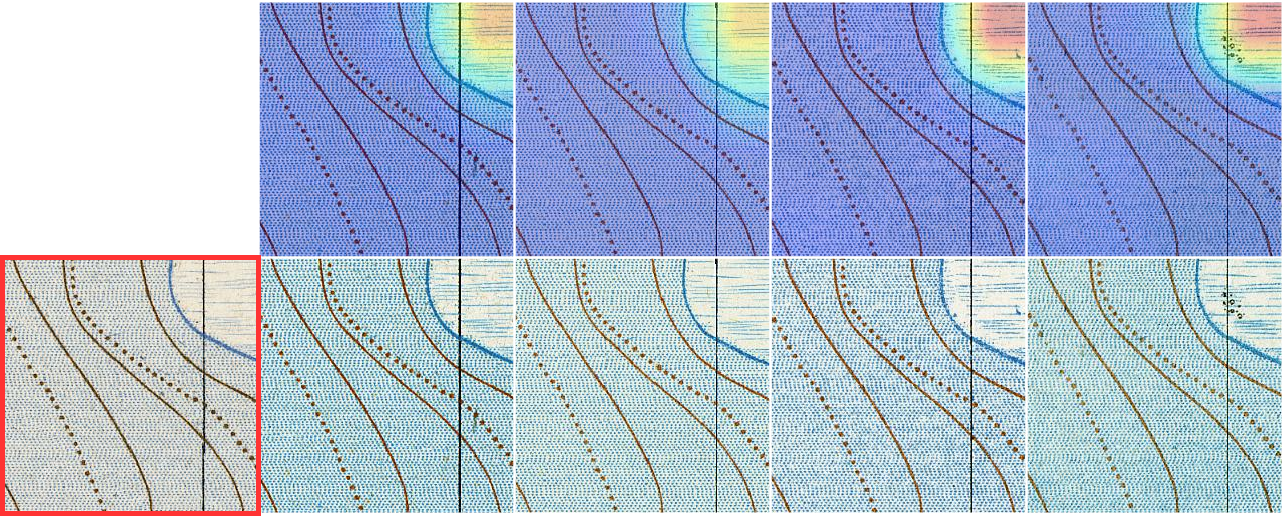}\\
\includegraphics[width=\linewidth]{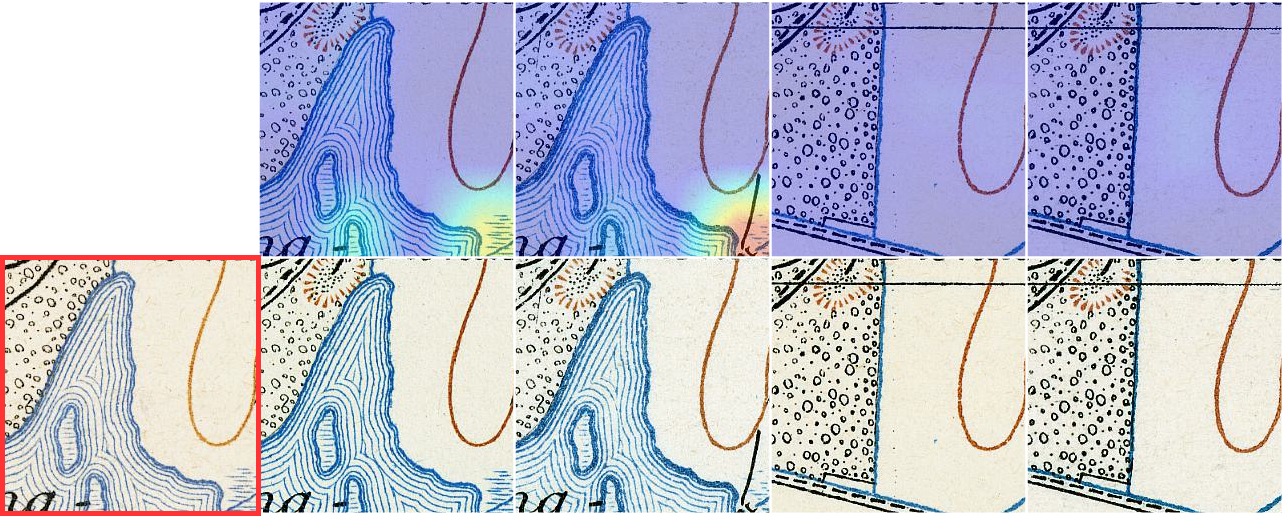}
\end{tabular}
}
\caption{Visualization of temporal attention. Blue to red represents values from low to high. We visualize the maximum value among all the attention heads. The center tile is marked as a red square and the tiles are connected chronologically from the left to right.}
\label{fig:temporalattention}
\end{figure*}
\end{sloppypar}
\clearpage

\bibliographystyle{ACM-Reference-Format}
\bibliography{references}
\end{document}